\documentclass[10pt,twocolumn,letterpaper]{article}

\usepackage[pagenumbers]{cvpr} 
\usepackage{tikz}
\usetikzlibrary{positioning} 
\usetikzlibrary{arrows.meta} 

\definecolor{cvprblue}{rgb}{0.21,0.49,0.74}
\usepackage[pagebackref,breaklinks,colorlinks,allcolors=cvprblue]{hyperref}

\def\paperID{*****} 
\def\confName{CVPR}
\def\confYear{2026}

\title{FecalFed: Privacy-Preserving Poultry Disease Detection via Federated Learning}

\author{Tien-Yu Chi\\
Independent Researcher\\
{\tt\small b03902059@ntu.edu.tw}
}

\begin{document}
\maketitle
\begin{abstract}
Early detection of highly pathogenic avian influenza (HPAI) and endemic poultry diseases is critical for global food security. While computer vision models excel at classifying diseases from fecal imaging, deploying these systems at scale is bottlenecked by farm data privacy concerns and institutional data silos. Furthermore, existing open-source agricultural datasets frequently suffer from severe, undocumented data contamination. In this paper, we introduce \textbf{FecalFed}, a privacy-preserving federated learning framework for poultry disease classification. We first curate and release \texttt{poultry-fecal-fl}, a rigorously deduplicated dataset of 8,770 unique images across four disease classes, revealing and eliminating a 46.89\% duplication rate in popular public repositories. To simulate realistic agricultural environments, we evaluate FecalFed under highly heterogeneous, non-IID conditions (Dirichlet $\alpha=0.5$). While isolated single-farm training collapses under this data heterogeneity, yielding only 64.86\% accuracy, our federated approach recovers performance without centralizing sensitive data. Specifically, utilizing server-side adaptive optimization (FedAdam) with a Swin-Small architecture achieves 90.31\% accuracy, closely approaching the centralized upper bound of 95.10\%. Furthermore, we demonstrate that an edge-optimized Swin-Tiny model maintains highly competitive performance at 89.74\%, establishing a highly efficient, privacy-first blueprint for on-farm avian disease monitoring.
\end{abstract}    
\section{Introduction}
\label{sec:intro}

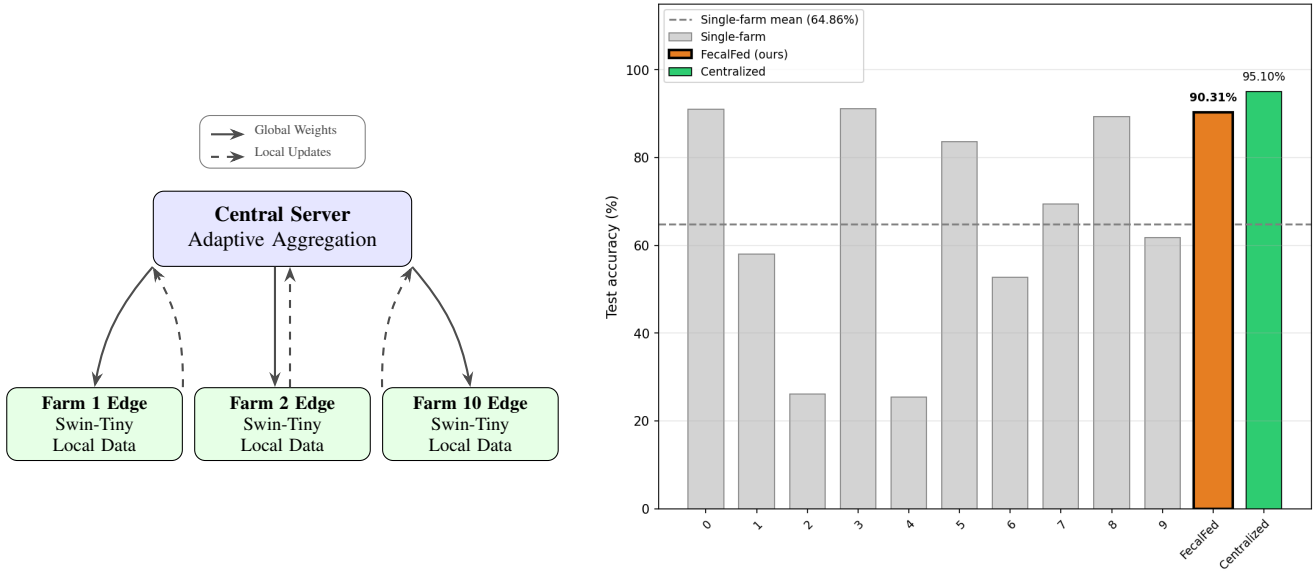
\begin{figure*}[t]
\centering

\begin{minipage}{0.38\textwidth} %
\centering
\begin{tikzpicture}[
    server/.style={rectangle, draw, fill=blue!10, text width=3.2cm, align=center, rounded corners, minimum height=1cm, font=\footnotesize},
    client/.style={rectangle, draw, fill=green!10, text width=2.1cm, align=center, rounded corners, minimum height=0.9cm, font=\scriptsize},
    arrow/.style={-{Stealth[length=2mm]}, thick, draw=black!70},
    dashed_arrow/.style={dashed, -{Stealth[length=2mm]}, thick, draw=black!70}
]

\node[server] (server) {\textbf{Central Server}\\Adaptive Aggregation};

\node[client, below=1.6cm of server] (farm2) {\textbf{Farm 2 Edge}\\Swin-Tiny\\Local Data};
\node[client, left=0.15cm of farm2] (farm1) {\textbf{Farm 1 Edge}\\Swin-Tiny\\Local Data};
\node[client, right=0.15cm of farm2] (farmN) {\textbf{Farm 10 Edge}\\Swin-Tiny\\Local Data};

\draw[arrow] (server.south west) to[bend right=15] (farm1.north);
\draw[dashed_arrow] (farm1.north east) to[bend right=15] (server.south west);

\draw[arrow] ([xshift=-0.1cm]server.south) -- ([xshift=-0.1cm]farm2.north);
\draw[dashed_arrow] ([xshift=0.1cm]farm2.north) -- ([xshift=0.1cm]server.south);

\draw[arrow] (server.south east) to[bend left=15] (farmN.north);
\draw[dashed_arrow] (farmN.north west) to[bend left=15] (server.south east);

\begin{scope}[shift={([xshift=-1.1cm, yshift=0.3cm]server.north)}]
    \draw[rounded corners, draw=black!50, fill=white] (0,0) rectangle (2.2, 0.7);
    
    \draw[arrow] (0.15, 0.45) -- (0.6, 0.45) node[right, font=\tiny, text=black!70, yshift=0.05cm] {Global Weights};
    
    \draw[dashed_arrow] (0.15, 0.15) -- (0.6, 0.15) node[right, font=\tiny, text=black!70, yshift=0.05cm] {Local Updates};
\end{scope}

\end{tikzpicture}
\end{minipage}%
\hfill
\begin{minipage}{0.55\textwidth} 
\centering

\includegraphics[width=\linewidth]{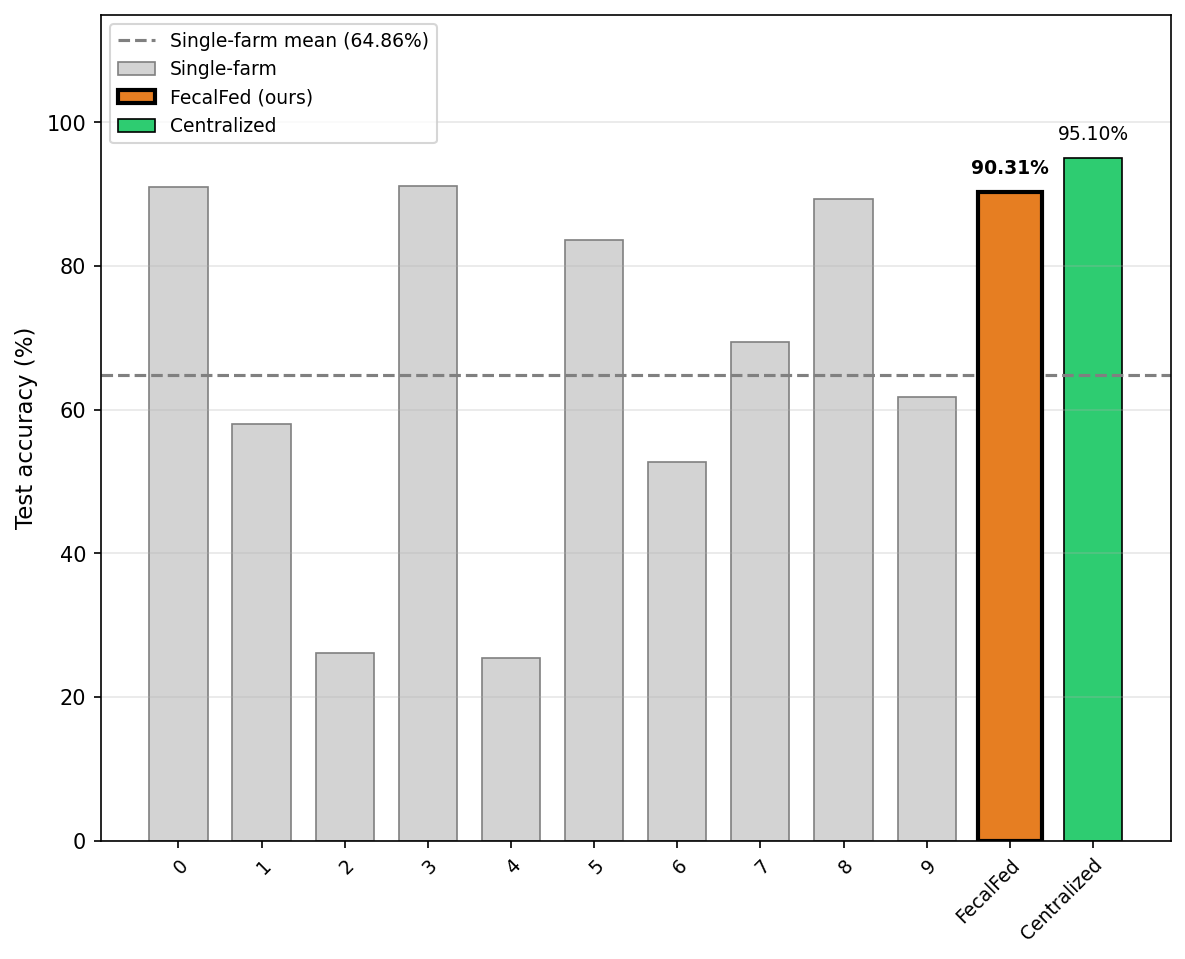}

\end{minipage}

\caption{Left: The FecalFed cross-silo federated learning architecture. Raw images remain isolated on-farm, while only Swin-Tiny model weights are communicated to the central server for FedAdam aggregation. Right: Test accuracy across the 10 isolated farm partitions (gray) demonstrating extreme variance and collapse under non-IID conditions, compared to the stable performance recovery achieved by FecalFed (orange).}
\label{fig:fecalfed_framework}
\end{figure*}

Highly pathogenic avian influenza (HPAI) continues to cause repeated global outbreaks, posing severe risks to agricultural economies, global food security, and public health \cite{FAO2025_H5N1_stepped_up_action, FAO_WHO_WOAH_2025_H5_joint_assessment, alders2025global}. Alongside HPAI, endemic infections such as Coccidiosis, Newcastle Disease (NCD), and Salmonella routinely threaten poultry flock health and lower industry productivity \cite{ahmad2023management, ganar2014newcastle, shaji2023salmonella}. Early detection is critical for preventing widespread outbreaks; however, traditional disease surveillance relies heavily on time-consuming, expensive laboratory procedures such as polymerase chain reaction (PCR) testing \cite{azeem2025diagnostic}. 

In recent years, deep learning and computer vision have emerged as powerful, non-invasive diagnostic alternatives \cite{degu2023smartphone, tasdelen2025detection, luong2024improving}. Fecal characteristics analysis is a proven visual inspection method, as gastrointestinal diseases drastically alter the shape, texture, and color of poultry droppings \cite{mozuriene2024physical}. 

Modern vision transformers have demonstrated high accuracy in classifying these diseases directly from fecal images, allowing for rapid, automated intervention without relying on expensive veterinary expertise \cite{liu2021swin, dosovitskiy2020image}.

Despite these algorithmic advances, deploying these systems at a global scale is severely bottlenecked by a dual challenge: institutional data silos and severe on-farm hardware limitations \cite{dembani2025agricultural}. Training robust artificial intelligence models typically requires aggregating massive datasets, but centralized training requires farms to upload raw images to a central server. Consequently, farms are reluctant to share sensitive health data due to biosecurity risks, commercial interests, and the fear of negative reputation impacts \cite{durrant2022role}. Furthermore, even if data privacy were not a concern, rural agricultural facilities frequently lack the enterprise-grade compute infrastructure and high-bandwidth internet required to support massive centralized models. As our experiments demonstrate, relying on isolated, single-farm models to bypass these data-sharing and hardware constraints is highly ineffective; under realistic heterogeneous data distributions, single-farm models collapse to a mean accuracy of 64.86\% from 95.1\%.

Furthermore, relying on existing open-source agricultural datasets to circumvent privacy restrictions introduces a hidden crisis: severe data contamination. Through rigorous perceptual hashing, we discovered that public repositories for poultry fecal images are heavily fragmented and overlapping. Our analysis of over 16,000 publicly available images revealed an alarming 46.89\% deduplication rate, with 77.4\% of images in popular synthetic datasets being mere resized duplicates of original field data \cite{fecal-lbh0j_dataset, disease-images-fecal_dataset, machuve_2021_5801834}. Training on such contaminated data leads to severe train/test leakage and artificially inflated performance metrics.

To overcome these dual challenges of data privacy and dataset quality, we introduce \textbf{FecalFed} (illustrated in Figure \ref{fig:fecalfed_framework}), a privacy-preserving federated learning (FL) framework for poultry disease monitoring, implemented using the open-source Flower (\texttt{flwr}) framework \cite{beutel2020flower}. 

In this work, we make the following core contributions:
\begin{itemize}
    \item \textbf{Curated Benchmark Dataset:} We release \texttt{poultry-fecal-fl}, a rigorously deduplicated, 4-class dataset comprising 8,770 unique images, establishing a clean, leak-free baseline for the agricultural AI community.
    \item \textbf{Realistic Non-IID Evaluation:} We demonstrate that isolated single-farm training fails to generalize under natural data heterogeneity, proving the absolute necessity of collaborative federated frameworks for robust disease detection.
    \item \textbf{Comprehensive Federated Benchmarking:} We systematically evaluate federated learning algorithms across different vision transformer architectures under non-IID conditions. We demonstrate that standard federated averaging (FedAvg) establishes a highly competitive baseline, while adaptive server-side optimization (FedAdam) provides further stability for larger models, achieving 90.31\% accuracy with a Swin-Small architecture.
\end{itemize}
\section{Related Work}
\label{sec:related}

Recent advancements in computer vision have facilitated automated poultry disease diagnostics through the analysis of fecal characteristics. Degu et al. \cite{degu2023smartphone} demonstrated the feasibility of smartphone-based convolutional neural networks (CNNs) for farm-level deployment, while Tasdelen et al. \cite{tasdelen2025detection} utilized transfer learning to improve CNN classification of high-risk diseases. Further pushing the boundaries of accuracy, Luong et al. \cite{luong2024improving} applied Vision Transformers combined with Integrated Gradients to provide explainable diagnostic outputs. However, these state-of-the-art approaches fundamentally rely on centralized data pools, entirely neglecting the restrictive data-sharing realities and privacy constraints of the modern agricultural sector.

The reluctance of farmers to share sensitive operational data due to biosecurity, competitive, and cybersecurity concerns is a well-documented barrier to centralized AI development \cite{kaur2022protecting, russell2025grow, wolfert2024farm}. Federated learning directly addresses this by enabling decentralized model training \cite{mcmahan2017communication}, a paradigm increasingly recognized for its potential to facilitate cross-silo collaboration in the agri-food sector \cite{durrant2022role, dembani2025agricultural}. While standard aggregation algorithms like FedAvg \cite{mcmahan2017communication} perform well under independent and identically distributed (IID) conditions, heterogeneous farm networks often require regularization techniques such as FedProx \cite{li2020federated} to restrict local model drift. However, because proximal terms can disrupt the fine-tuning of large pre-trained vision models, adaptive server-side optimizers like FedAdam \cite{reddi2020adaptive} present a crucial but currently underexplored alternative for stabilizing non-IID agricultural edge deployments.

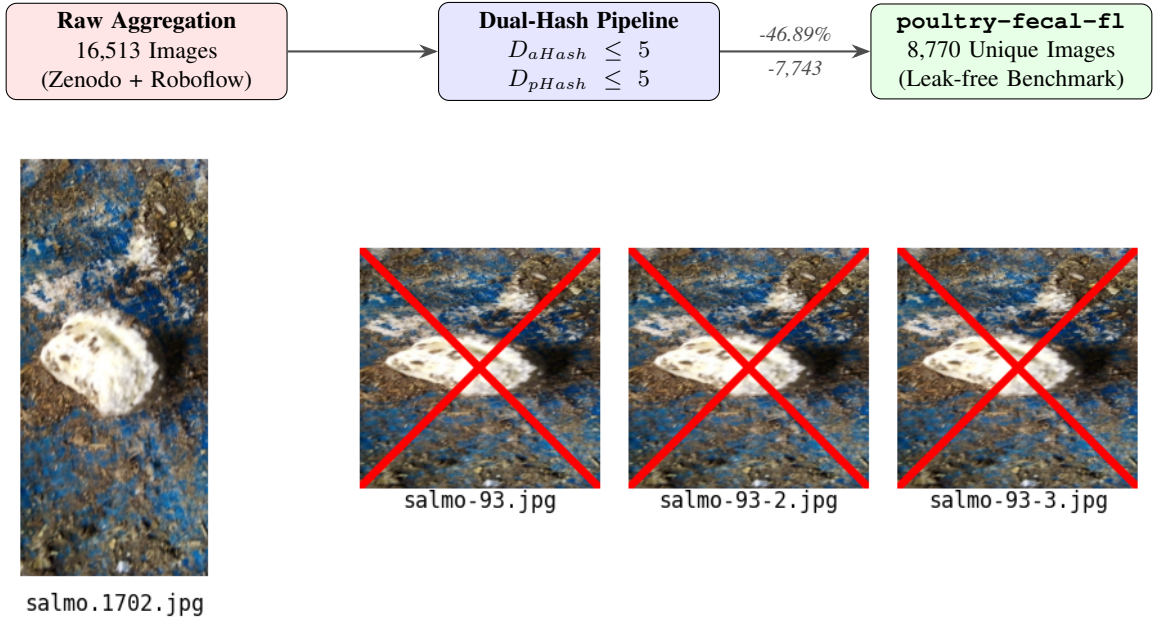
\begin{figure*}[t]
\centering
\begin{tikzpicture}[
    node distance=1.5cm and 2cm,
    box/.style={rectangle, draw, fill=blue!10, text width=3.5cm, align=center, rounded corners, minimum height=1.2cm, font=\small},
    arrow/.style={-{Stealth[length=2.5mm]}, thick, draw=black!70},
    label/.style={font=\footnotesize\itshape, text=black!70}
]

\node[box, fill=red!10] (raw) {\textbf{Raw Aggregation}\\16,513 Images\\(Zenodo + Roboflow)};
\node[box, right=of raw] (hash) {\textbf{Dual-Hash Pipeline}\\$D_{aHash} \le 5$\\$D_{pHash} \le 5$};
\node[box, fill=green!10, right=of hash] (clean) {\textbf{\texttt{poultry-fecal-fl}}\\8,770 Unique Images\\(Leak-free Benchmark)};

\draw[arrow] (raw) -- node[above, label] {} (hash);
\draw[arrow] (hash) -- node[above, label] {-46.89\%} node[below, label] {-7,743} (clean);

\end{tikzpicture}

\vspace{0.5cm} 

\includegraphics[width=0.88\linewidth]{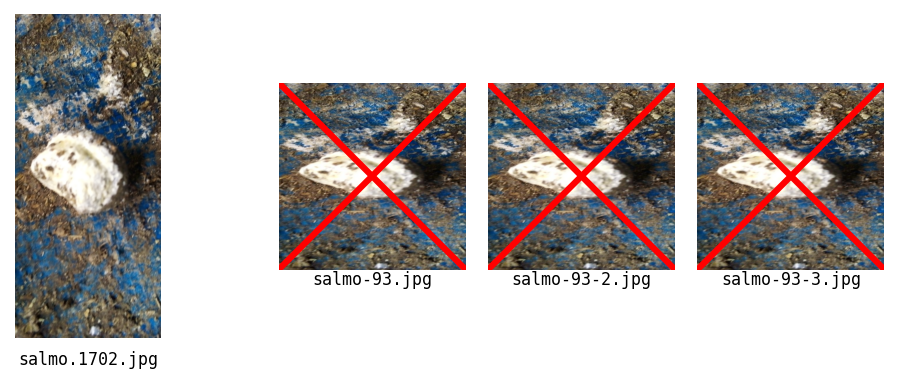}

\caption{The FecalFed data curation pipeline. (Top) The dual-hash deduplication process eliminated 46.89\% of the raw aggregated dataset. (Bottom) A visual example of severe cross-source contamination, where original field data was found repeatedly down-sampled and duplicated in synthetic open-source repositories.}
\label{fig:data_leakage}
\end{figure*}

Finally, the transition from centralized to federated agricultural AI is complicated by a growing reproducibility crisis driven by data leakage in public repositories \cite{kapoor2023leakage}. In computer vision, unchecked cross-source duplication artificially inflates performance metrics and severely undermines the reliability of visual datasets \cite{ramos2025data}. To combat this, automated near-duplicate image detection utilizing perceptual hashing and deep learning has become an essential prerequisite for dataset sanitization \cite{jakhar2025effective}. Because existing poultry disease repositories lack these rigorous deduplication protocols, establishing a sanitized, farm-split benchmark is a necessary foundation for evaluating federated agricultural frameworks.
\section{Data Curation and Preprocessing}
\label{sec:data}

A primary barrier to robust agricultural AI is the fragmentation and undocumented contamination of public datasets. To establish a rigorous baseline for our federated learning experiments, we developed a comprehensive data curation pipeline to aggregate, sanitize, and standardize poultry fecal imagery from multiple open-source repositories. 

\subsection{Raw Data Acquisition and Standardization}
We constructed our initial data pool by aggregating multi-class fecal image datasets from two primary sources: the Zenodo AI4D Tanzania collection \cite{machuve_2021_5801834} and two public Roboflow Universe repositories (\textit{disease-images-fecal} \cite{disease-images-fecal_dataset} and \textit{fecal-lbh0j} \cite{fecal-lbh0j_dataset}). The raw aggregation yielded 16,513 images categorized into four diagnostic classes: Healthy, Coccidiosis, Newcastle Disease (NCD), and Salmonella.

Because the \textit{fecal-lbh0j} dataset was originally formatted for object detection (COCO format), we developed a conversion script to extract bounding boxes and reformat them for image-level classification. To ensure strict single-label classification clarity, any image containing bounding boxes for multiple different disease classes was entirely discarded from the dataset. 

\subsection{Data Leakage and Deduplication Pipeline}
Merging independent public repositories introduces a high risk of data contamination, as augmented or resized versions of identical images are frequently re-uploaded across different platforms. To prevent train-test leakage and artificial inflation of our federated model performance, we implemented a dual-algorithm perceptual hashing deduplication pipeline. 

For every image, we computed a 256-bit Average Hash (aHash) to capture macro-level brightness patterns, alongside a Perceptual Hash (pHash) utilizing Discrete Cosine Transforms (DCT) to capture frequency-domain features. Image pairs were flagged as duplicates and removed if the Hamming distance for both hashes was less than or equal to a strict threshold of 5:
\begin{equation}
    D_{aHash}(x, y) \le 5 \quad \land \quad D_{pHash}(x, y) \le 5
\end{equation}

Our analysis revealed an alarming degree of data contamination within the open-source agricultural community. As illustrated in Figure \ref{fig:data_leakage}, of the 16,513 raw images processed, 7,743 were identified as duplicates, resulting in a 46.89\% dataset reduction. Most notably, we discovered that 77.4\% of the Roboflow images were merely down-sampled duplicates of the original high-resolution Zenodo field data. Furthermore, we eliminated 19 cross-label duplicate groups where identical images were assigned contradictory disease labels in different repositories. The final, meticulously sanitized \texttt{poultry-fecal-fl} dataset comprises 8,770 unique images.

\subsection{Image Preprocessing and Augmentation}
To prepare the deduplicated dataset for Vision Transformer architectures, all images were standardized to a resolution of $224 \times 224$ pixels. To improve model generalization and simulate varied on-farm lighting conditions, we applied a robust stochastic data augmentation pipeline during training. This included random resized cropping, random horizontal and vertical flips, and random rotations up to $30^\circ$. Additionally, color jitter was applied with a $0.2$ variance factor for brightness, contrast, and saturation, and $0.1$ for hue. Finally, all input tensors were normalized using standard ImageNet channel statistics.
\section{Federated Learning Methodology}
\label{sec:methodology}

To evaluate the feasibility of privacy-preserving disease detection, we formulated the classification task as a cross-silo federated learning problem. Rather than centralizing the $8,770$ deduplicated images, data remains localized on simulated farm edge devices, and only model weight updates are transmitted to a central aggregation server. All federated experiments were orchestrated using the open-source Flower (\texttt{flwr})\cite{beutel2020flower} framework.

\begin{figure}[t]
\centering

\includegraphics[width=\linewidth]{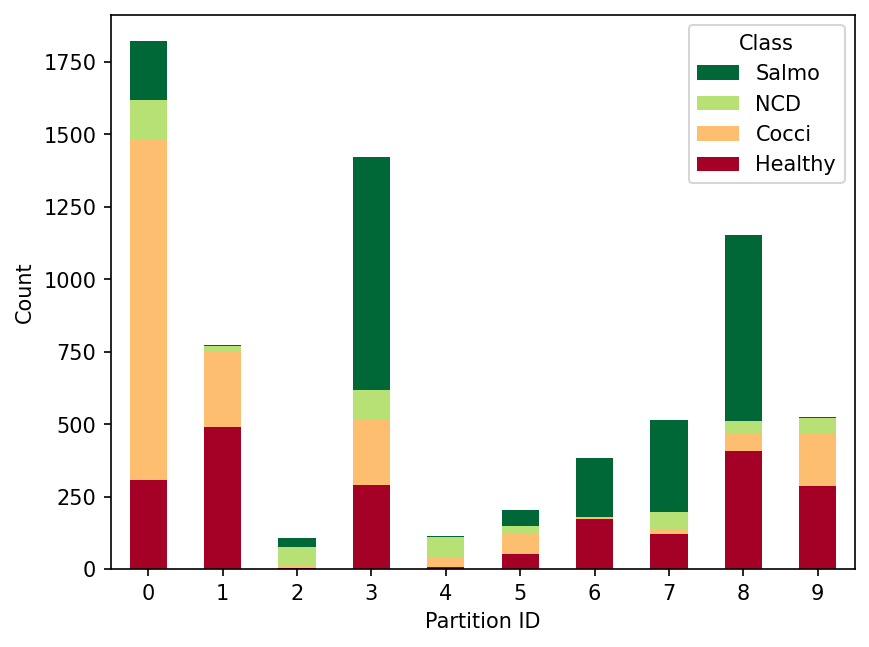}

\caption{Data distribution across 10 isolated farm clients under a Dirichlet ($\alpha=0.5$) non-IID partitioning strategy. This extreme skew mimics realistic agricultural settings where localized outbreaks cause specific edge devices to hold a vast majority of samples for a single disease.}
\label{fig:dirichlet_dist}
\end{figure}

\subsection{Realistic Non-IID Data Partitioning}
A major limitation of existing agricultural AI literature is the assumption of Independent and Identically Distributed (IID) data. In reality, poultry disease outbreaks are highly localized; one farm may experience a severe Coccidiosis outbreak while another remains perfectly healthy. 

To simulate this natural, heterogeneous environment across our network of $10$ isolated farms, we partitioned the \texttt{poultry-fecal-fl} dataset using a Dirichlet distribution. By setting the concentration parameter to $\alpha=0.5$, we forced highly skewed, non-IID label distributions across the clients. As visualized in Figure \ref{fig:dirichlet_dist} [Placeholder for Figure 3], some clients possess a majority of Salmonella samples, while others contain predominantly Healthy or NCD samples, perfectly mimicking the unpredictable nature of real-world farm data.

\subsection{Vision Architectures and Edge Deployment}

We evaluated four distinct Vision Foundation Model architectures to determine the optimal balance between diagnostic accuracy and edge-device deployability. A critical requirement of cross-silo federated learning is that participating farm clients must possess the memory and compute capacity to execute local model training, not merely inference. For our high-capacity baselines, we utilized the standard Vision Transformer (ViT-B/16, 86M parameters) \cite{dosovitskiy2020image} and the hierarchical Swin-Small transformer (50M parameters) \cite{liu2021swin}. To address the strict computational constraints of typical agricultural edge nodes, such as smartphone-based diagnostics or embedded IoT systems like the NVIDIA Jetson Nano, we concurrently evaluated two lightweight models: ViT-S/16 (22M parameters) and Swin-Tiny (28M parameters). For all architectures, the pre-trained feature extraction backbones were frozen, and only the classification heads were fine-tuned during the federated rounds. This strategy deliberately minimizes both communication overhead and the local device memory footprint, making on-farm training highly feasible.

\subsection{Optimization Strategies and Hyperparameters}
In each federated communication round, the central server randomly sampled $50\%$ of the network ($5$ out of $10$ farms) to participate in training. Selected clients trained their local models for $E=1$ local epoch utilizing a batch size of $256$ simulated on an NVIDIA A100 GPU. 

\begin{table*}[t]
\centering
\caption{Performance comparison of vision foundation models across training paradigms under non-IID conditions (Dirichlet $\alpha=0.5$). The baseline for isolated single-farm training collapses to the $60\%$--$65\%$ range across all architectures due to extreme local data skew. Federated learning successfully recovers this lost performance, with FedAdam providing the highest peak accuracy for larger models and minimizing the gap to the centralized upper bound.}
\label{tab:fl_results}
\begin{tabular}{lcccccc}
\toprule
\textbf{\begin{tabular}{@{}l@{}}Model \\ Architecture\end{tabular}} & 
\textbf{Parameters} & 
\textbf{\begin{tabular}{@{}c@{}}Single-Farm \\ (Non-IID)\end{tabular}} & 
\textbf{\begin{tabular}{@{}c@{}}Centralized \\ (Upper Bound)\end{tabular}} & 
\textbf{\begin{tabular}{@{}c@{}}FL \\ (FedAvg)\end{tabular}} & 
\textbf{\begin{tabular}{@{}c@{}}FL \\ (FedAdam)\end{tabular}} & 
\textbf{\begin{tabular}{@{}c@{}}Gap (Best FL \\ vs. Cent.)\end{tabular}} \\
\midrule
Swin-Small & 50M & 64.86\% $\pm$ 24.95\% & 95.10\% & 89.74\% & \textbf{90.31\%} & -4.79\% \\
ViT-B/16 & 86M & 60.97\% $\pm$ 23.05\% & 94.81\% & 88.77\% & 90.02\% & -4.79\% \\
Swin-Tiny & 28M & 64.03\% $\pm$ 24.04\% & 93.04\% & 86.89\% & 89.74\% & \textbf{-3.30\%} \\
ViT-S/16 & 22M & 65.87\% $\pm$ 22.06\% & 92.99\% & \textbf{89.28\%} & 85.12\% & -3.71\% \\
\bottomrule
\end{tabular}
\end{table*}

We benchmarked two primary server-side aggregation strategies:
\begin{itemize}
    \item \textbf{Federated Averaging (FedAvg):} The standard baseline, which computes a strict, data-weighted average of the local client updates.
    \item \textbf{Adaptive Server Optimization (FedAdam):} Because heterogeneous (non-IID) client updates can cause standard aggregation to diverge or stall—particularly for large vision models—we applied FedAdam. This strategy treats the aggregated pseudo-gradient as an input to a server-side Adam optimizer. 
\end{itemize}

For our optimal FedAdam configuration, the server-side hyperparameters were explicitly set to a learning rate of $\eta=0.1$, moment decays of $\beta_1=0.9$ and $\beta_2=0.99$, and a degree of adaptability $\tau=0.001$. All federated training sessions were executed for $10$ global communication rounds, with global evaluation performed centrally at the end of each round to track accuracy and loss convergence.
\section{Experiments and Results}
\label{sec:experiments}

To rigorously evaluate our FecalFed framework, we benchmarked multiple vision models under centralized, isolated, and federated paradigms. All federated simulations utilized the \texttt{poultry-fecal-fl} dataset distributed across 10 clients.

\subsection{Establishing Baselines: Centralized vs. Isolated}
We first established an upper-bound performance ceiling by training our models centrally on the fully aggregated dataset. Under these ideal conditions, the Swin-Small architecture achieved a peak accuracy of 95.10\%, while the lightweight Swin-Tiny achieved 93.04\%.

Conversely, to establish the lower bound and demonstrate the vulnerability of isolated farm networks, we evaluated single-client training. Under realistic non-IID conditions (Dirichlet $\alpha=0.5$), performance collapsed entirely. The mean single-farm accuracy plummeted to 64.86\% with extreme variance ($\pm 24.95\%$), as models catastrophically overfit to local disease distributions. This 30.24\% drop from the centralized baseline proves that isolated training is fundamentally unviable for robust agricultural diagnostics.

\subsection{Federated Model Benchmarking}
To recover this lost performance without centralizing raw data, we evaluated standard Federated Averaging (FedAvg) and adaptive server-side optimization (FedAdam) across our vision models. As shown in Table \ref{tab:fl_results}, FecalFed successfully bridges the gap to the centralized upper bound. 

Utilizing FedAdam ($\eta=0.1$) with the Swin-Small architecture yielded our best federated result, achieving 90.31\% accuracy. This represents a massive +25.45\% absolute improvement over isolated training, closing the gap to the centralized baseline to a mere 4.79\%. 

\begin{figure}[t]
\centering

\includegraphics[width=\linewidth]{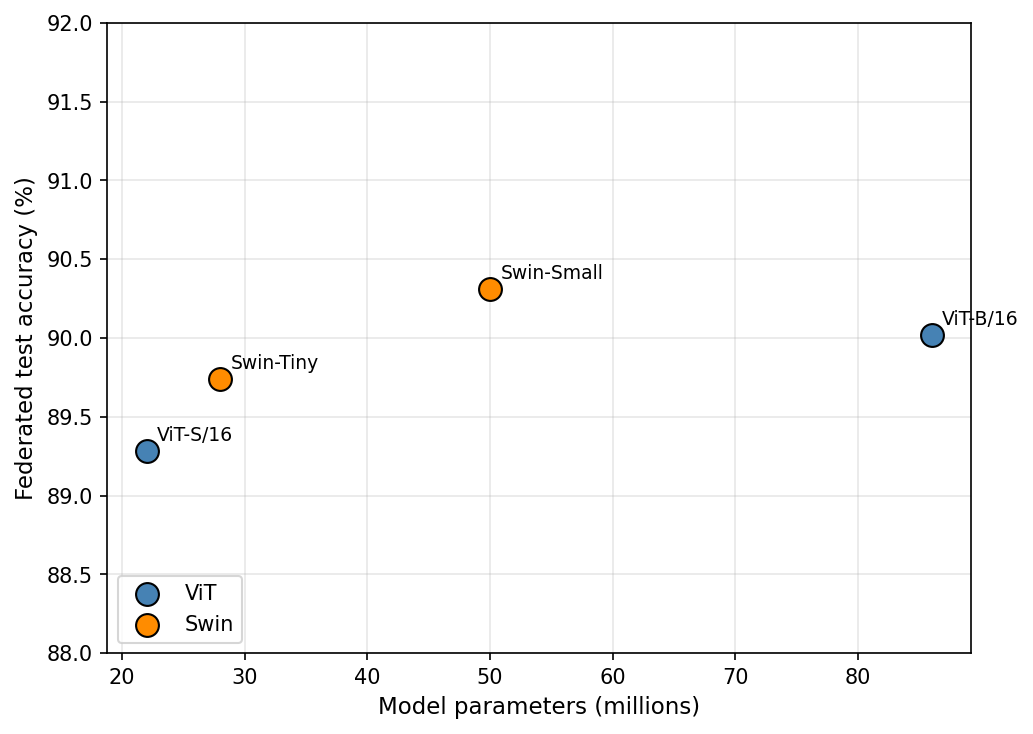}

\caption{Federated test accuracy versus model parameter footprint. The Swin-Tiny architecture (28M parameters) emerges as the optimal candidate for edge deployment, achieving highly competitive accuracy (89.74\%) while drastically reducing the computational and memory requirements compared to larger models like ViT-B/16.}
\label{fig:param_efficiency}
\end{figure}

Crucially for edge deployments, we found that the Swin-Tiny model (28M parameters) maintained a highly competitive 89.74\% accuracy under FedAdam. By achieving an accuracy within 0.28\% of the much larger ViT-B/16 baseline (90.02\%) while utilizing less than a third of its parameter footprint (28M vs. 86M), Swin-Tiny establishes an optimal operational sweet spot. It provides high-fidelity diagnostic performance while remaining small enough to fit within the memory and compute constraints of standard agricultural edge computing nodes.

\subsection{Ablation Study: Communication Efficiency and Convergence}

To evaluate model convergence and the impact of extended training on non-IID data distributions, we conducted an ablation study varying the total number of global communication rounds. Using the ViT-B/16 architecture under the standard FedAvg strategy, we scaled the training duration from 5 up to 20 rounds. 

\begin{figure}[t]
\centering
\includegraphics[width=\linewidth]{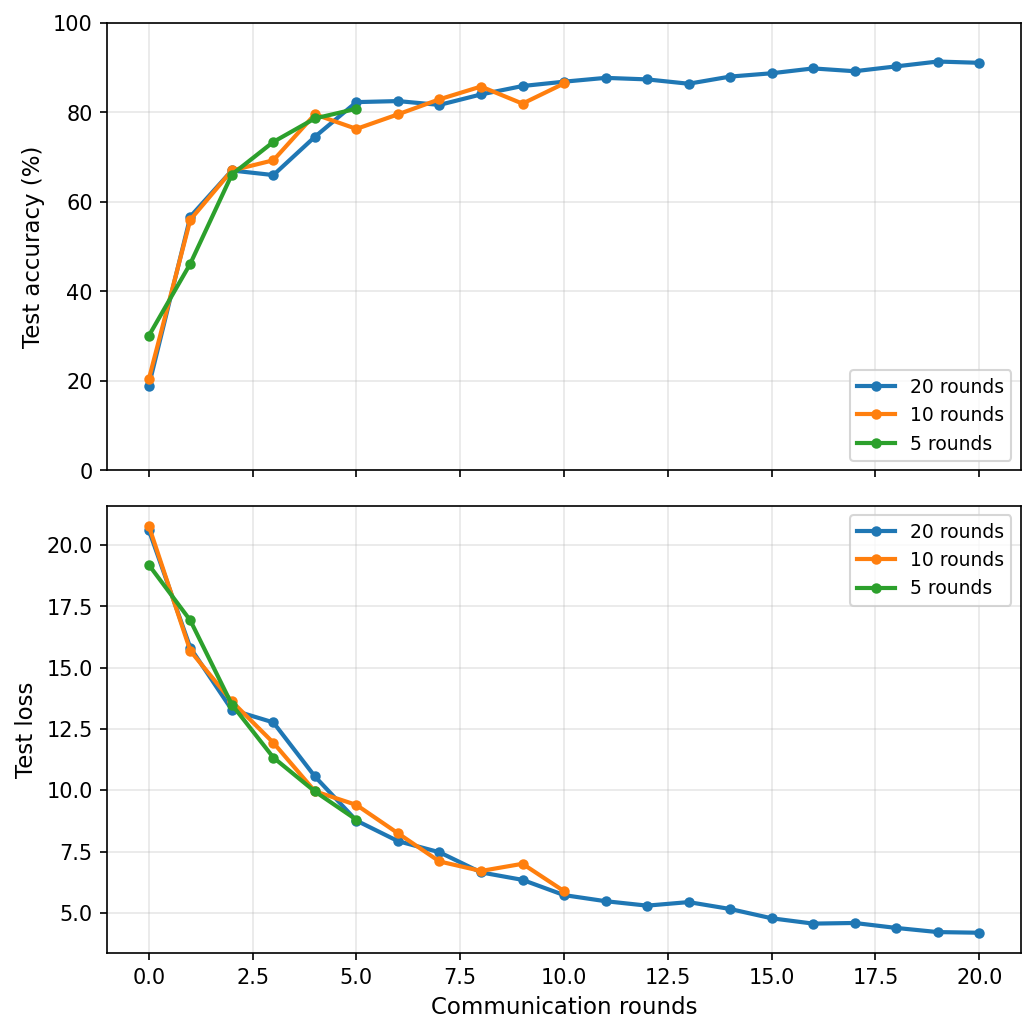}
\caption{Ablation study on global communication rounds using the ViT-B/16 baseline under FedAvg. The top panel shows test accuracy convergence, while the bottom panel shows test loss decay. Extending the training duration to 20 rounds continuously mitigates the effects of non-IID data skew, yielding smooth, stable performance improvements without stalling.}
\label{fig:ablation_convergence}
\end{figure}

As visualized in Figure \ref{fig:ablation_convergence}, extending the federated training consistently mitigated the effects of local data skew. The convergence curves demonstrate that longer training durations provide smooth, continuous improvements in both test loss and accuracy, ultimately elevating performance from 80.79\% at 5 rounds to a peak of 91.05\% at 20 rounds. Crucially, because FecalFed operates on frozen vision model backbones and strictly aggregates the lightweight weights of the fine-tuned classification heads, these extended communication rounds incur minimal bandwidth overhead. This high communication efficiency ensures that achieving optimal model convergence remains highly practical, even for real-world agricultural deployments constrained by unstable or low-bandwidth internet connectivity.
\section{Conclusion}
\label{sec:conclusion}

In this paper, we introduced FecalFed, a privacy-preserving federated learning framework designed to overcome the critical data silos and biosecurity constraints hindering agricultural AI. Recognizing that robust federated evaluation requires uncontaminated baselines, we first curated and released \texttt{poultry-fecal-fl}, exposing and eliminating a severe 46.89\% duplication rate in widely used public repositories. Through extensive non-IID benchmarking, we demonstrated that isolated, single-farm training collapses under natural data heterogeneity. By implementing adaptive server-side optimization (FedAdam), FecalFed successfully recovers this performance, enabling a Swin-Small vision foundation model to achieve 90.31\% accuracy without ever centralizing sensitive raw data. Furthermore, we established a highly practical blueprint for edge deployment, demonstrating that the lightweight Swin-Tiny architecture maintains competitive diagnostic performance (89.74\%) while satisfying the strict computational constraints of on-farm embedded systems. Ultimately, this work provides both the sanitized data baseline and the decentralized training framework necessary for scalable, global poultry disease surveillance.
{
    \small
    \bibliographystyle{ieeenat_fullname}
    \bibliography{main}
}


\end{document}